# Kriformer: A Novel Spatiotemporal Kriging Approach Based on Graph Transformers

**Renbin Pan [1], Feng Xiao [2,*], Hegui Zhang [3], Minyu Shen [4]**

[1] School of Business Administration, Southwestern University of Finance and Economics, Chengdu 611130, China

[2,*] Business School, Sichuan University, Chengdu 610064, China, corresponding author, Email: evan.fxiao@gmail.com

[3] School of Data Science and Artificial Intelligence, Dongbei University of Finance and Economics, Dalian 116025, China

[4] School of Management Science and Engineering, Southwestern University of Finance and Economics, China, Chengdu 611130, China

**Abstract:**

Accurately estimating data in sensor-less areas is crucial for understanding system dynamics, such as traffic state estimation and environmental monitoring. This study addresses challenges posed by sparse sensor deployment and unreliable data by framing the problem as a spatiotemporal kriging task and proposing a novel graph transformer model, Kriformer. This model estimates data at locations without sensors by mining spatial and temporal correlations, even with limited resources. Kriformer utilizes transformer architecture to enhance the model's perceptual range and solve edge information aggregation challenges, capturing spatiotemporal information effectively. A carefully constructed positional encoding module embeds the spatiotemporal features of nodes, while a sophisticated spatiotemporal attention mechanism enhances estimation accuracy. The multi-head spatial interaction attention module captures subtle spatial relationships between observed and unobserved locations. During training, a random masking strategy prompts the model to learn with partial information loss, allowing the spatiotemporal embedding and multi-head attention mechanisms to synergistically capture correlations among locations. Experimental results show that Kriformer excels in representation learning for unobserved locations, validated on two real-world traffic speed datasets, demonstrating its effectiveness in spatiotemporal kriging tasks.

## 1. Introduction
### 1.1 Background

Estimating spatially and temporally correlated variables across a network with precision is crucial in numerous fields, including traffic speed monitoring (Liang et al., 2018; Tu et al., 2021) and air pollution assessment (Nebenzal et al., 2020). Nonetheless, gaining a thorough comprehension of the variable throughout the entire network presents a formidable challenge. This difficulty arises from the substantial expenses involved in deploying and maintaining sensors, which hinder the establishment of a comprehensive sensing infrastructure. Moreover, the periodic malfunctioning of sensors can exacerbate the issue by causing data gaps, further compromising the accuracy and reliability of network-wide sensing.

Fortunately, many variables within networks exhibit significant spatial and temporal correlations. These correlations can be leveraged to reconstruct undetected or inaccurately measured values on specific network components (Nie et al., 2023). Recent research has primarily focused on addressing the spatiotemporal kriging problem, which aims to estimate the values of a variable on unobserved components within a network. Unlike traditional prediction tasks, spatiotemporal kriging does not rely on historical information to estimate future values. Instead, it utilizes measurements from other observed components to perform "spatial prediction." By effectively harnessing the intrinsic multi-dimensional correlations, sparse sensors can achieve a "collaborative perception" of the variable on undetected network connections. Figure 1 illustrates an example in traffic data interpolation, where, by considering spatiotemporal correlations, the data of undetected road 3 can potentially be inferred from other roads, which have sensors installed.

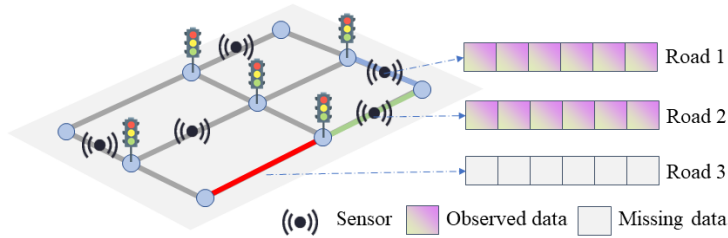

(a) Examples of spatial proximity

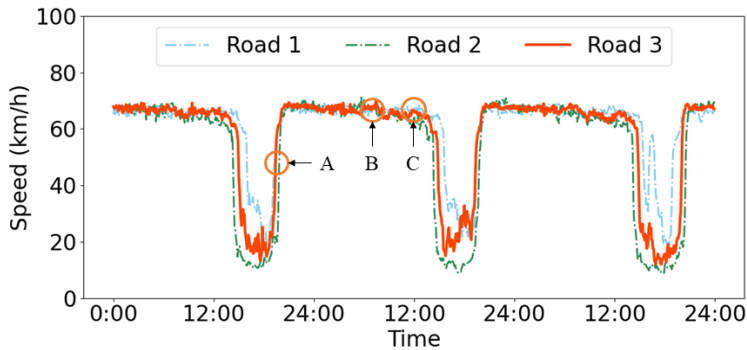

(b) Examples of diverse temporal patterns

Figure 1: Illustration of the spatiotemporal kriging problem using traffic speed dataset in Los Angeles. Roads 1 and 2 are equipped with sensors recording traffic speeds, while road 3 lacks sensor data, resulting in the absence of historical records. The aim of spatiotemporal kriging is to estimate traffic speeds at undetected locations (road 3) by utilizing multi-dimensional spatiotemporal correlations observed at roads 1 and 2. In Figure 1(a), road 2 demonstrates closer physical proximity to road 3 than to road 1, suggesting lower similarity errors, such as Mean Squared Error (MSE), between road 2 and road 3 compared to those between road 1 and road 3. In Figure 1(b), the traffic speed exhibits periodicity and variability, and estimation results for nearby temporal positions within the time series are often more correlated. For example, due to their temporal proximity, the traffic speed at time point C is closer in value to the speed at time point B than to that at time point

A.

**1.2 Key challenges and research gaps**

The pivotal focus of the kriging problem lies in effectively harnessing correlations between undetected road nodes and other sparsely detected roads. Three primary challenges and related research gaps are discussed as follows.

Firstly, the correlation between observed and unobserved component values can be complex, extending beyond proximal locations. Even at greater distances, valuable information might be provided. To accurately capture correlations for each unobserved location, it is crucial to consider the intricate structural relationships between positions from a global attention perspective. Moreover, treating unobserved nodes required for kriging interpolation as virtual nodes has been shown to facilitate the propagation of global information across the graph (Gilmer et al., 2017; Ishiguro et al., 2019; Pham et al., 2017). However, certain existing literature employing graph-based approaches for kriging problems has solely focused on observed nodes within the graph, limiting the information propagation capability between observed and unobserved nodes on the graph.

Secondly, although some research has proposed Graph Neural Networks (GNN) methods based on message passing to estimate kriging and has achieved certain estimation effects, recent studies have highlighted limitations in GNN (Alon and Yahav, 2020; Chen et al., 2022; Ma et al., 2023; Zhang et al., 2021). GNN that solely consider local dependencies result in learned representations containing incomplete information. This limitation may further restrict model expressiveness, lead to excessive smoothing, and compress information. As GNN architectures deepen, models may become unresponsive to training data, and the representations obtained by such deep models often become overly smooth, causing node representations to converge and become difficult to distinguish.

Lastly, temporal correlation is crucial for kriging problems due to the time-varying, dynamic, periodic, and interdependent nature of the variable. However, previous research has not proposed effective ways to capture temporal correlations, particularly addressing long-range correlations, despite their significance for variations in the variable.

The transformer architecture and its derivatives, such as Bert and ViT, have achieved substantial success in natural language processing and computer vision (Devlin et al., 2018; Dosovitskiy et al., 2020; Vaswani et al., 2017). While demonstrating proficiency in predicting traffic state and demand (Guo et al., 2022; Wen et al., 2023; Yan et al., 2022; Yang et al., 2023), the kriging problem poses distinct challenges due to its lack of historical data, intensifying its complexity. Consequently, the direct application of transformers proposed by these studies to kriging problems is not feasible.

**1.3 Contributions**

To address the prevalent challenges in estimating data for undetected regions across various domains, this research introduces a novel kriging model, grounded in the graph transformer architecture. This model leverages the inherent multi-dimensional spatiotemporal correlations within the data to effectively tackle the estimation of information in areas lacking direct sensor measurements. The primary contributions of this study include:

(1) We have developed a novel framework, namely Kriformer, which, to the best of our knowledge, is the first method based on the transformer architecture used to address spatiotemporal kriging tasks. In this framework, we employ global nodes (observed and virtual nodes) and introduce a random masking strategy for iterative training. The incorporation of virtual nodes promotes the propagation of global information across the graph.

(2) Within the Kriformer framework, we introduced spatial node representations (Eigenmaps) combined with the temporal embedding approach, constructing the Spatiotemporal Embedding (STE) module. By integrating spatial and temporal information, this module not only builds richer feature representations and provides the model with more comprehensive contextual information, but also allows the Kriformer framework to dynamically adjust attention allocation strategies based on spatiotemporal embeddings. This enhances the model's ability to capture complex spatiotemporal patterns of traffic speeds among nodes, thereby improving its learning effectiveness.

(3) In the proposed Kriformer framework, we redesigned spatial and temporal attention mechanisms. Particularly, we introduced the Multi-head Spatial Interaction Attention (MSIA) module, aimed at capturing the attentional distribution relationships of traffic speed between observed and unobserved nodes to facilitate spatiotemporal feature learning. Simultaneously, we retained the original transformer architecture, utilizing an encoder-decoder structure to address the kriging interpolation problem.

(4) We conducted extensive experiments on two real traffic datasets, revealing that our model achieves superior kriging accuracy compared to baseline models. Additionally, we conducted ablation studies and sensitivity analyses of key hyperparameters, providing valuable insights for the practical application of our Kriformer model.

**1.4 Roadmap**
The remainder of this paper are organized as follows. Section 2 offers an overview of various kriging approaches and explores relevant literature on transformer models. Section 3 introduces the definition of the kriging problem. Section 4 details our proposed model framework, elucidating its components comprehensively. Section 5 delves into quantitative experiments, while Section 6 underscores notable conclusions.

**2. Literature Review**
Kriging, a spatial interpolation method, aims to estimate variable values within a continuous spatial field using limited data points (Stein, 1999). Recently, the kriging task has piqued substantial interest within various disciplines, often seen as specialized variants of kriging. This section delves into primary approaches addressing spatiotemporal kriging interpolation problems, categorizing them into two main classes: statistical models and deep learning models. Subsequently, we systematically classify and introduce the existing literature on these methods.

**2.1 Statistical Methods in Kriging**
The kriging interpolation task stands distinct from simple missing data imputation. While researchers have devised several statistical models (e.g., Tensor factorization (Chen et al., 2019)) for various missing data imputation tasks, they are relatively less intricate compared to the kriging interpolation challenge. Consequently, specific studies have tailored statistical models for kriging, such as Ordinary kriging and Universal kriging (Chilès and Desassis, 2018). However, these models, designed for different data sources, do not seamlessly apply to our problem domain, which revolves around traffic speed considerations.

Bae et al. (2018) proposed a cokriging method designed to capitalize on the spatiotemporal dependencies inherent in traffic data. This approach utilizes multiple data sources to compute high-resolution traffic speed. Although the results indicate that the spatiotemporal cokriging method with multiple data sources can effectively enhance completion accuracy, the outcomes of kriging may be affected when data from other sources are inaccurate. Lei et al. (2022) introduced a Bayesian kernel matrix decomposition model for solving spatiotemporal traffic kriging. This method employs a fully Bayesian approach by learning hyperparameters of graph kernels through resampling. An undeniable drawback is the time-consuming process of hyperparameter sampling, possibly presenting a trade-off between model scalability and flexibility. Nie et al. (2023) proposed a Laplacian-Enhanced Low-Rank Tensor Completion (LETC) framework with low-rank and multi-dimensional correlations for large-scale traffic speed kriging under finite observations. Despite achieving promising results, the framework requires careful selection of spatiotemporal data correlations and the simultaneous use of three different forms of graph Laplacian operators for modeling. This complexity may present challenges in understanding and implementing the model in practical applications.

**2.2 Deep Learning Approaches in Kriging**
Graph Neural Networks (GNN) have excelled in various graph representation learning scenarios and exhibited impressive performance in kriging tasks. Appleby et al. (2020) introduced a kriging convolutional network using k-nearest neighbors to reconstruct all node values within a graph. Wu et al. (2021) proposed an inductive GNN aimed at modeling spatial dependencies, employing a random mask sampling method for model training. However, these methods consider samples within specific observation times solely as individual or additional features, neglecting temporal correlations. Liang et al. (2022) elevated the kriging performance of GNN by integrating an external attention mechanism and temporal convolutional networks. While GNN have achieved state-of-the-art performance in kriging tasks, their restricted receptive fields have implications. Shallow GNN are limited to aggregating nearby information, resulting in pronounced structural biases and noise. Conversely, deep GNN tends to be over-smooth, aggregating a substantial amount of unrelated information (Alon and Yahav, 2020; Chen et al., 2022; Ma et al., 2023; Zhang and Xiao, 2021).

**2.3 Transformer Methods and Applications**
The transformer was initially proposed in the field of machine translation, with its core idea being the attention mechanism (Vaswani et al., 2017). Within the transformer architecture, self-attention plays a pivotal role, transforming sequences by computing context-aware vectors for symbols in the input sequence. This method ensures that each symbol's representation is influenced by all other symbols, establishing an extensive global receptive field. Consequently, self-attention serves as a highly adaptable mechanism capable of capturing intricate dynamics and long-term patterns within sequence data. Renowned for its efficacy in representation learning, it swiftly supplanted recurrent and convolutional neural networks, emerging as a premier model across diverse tasks (Guo et al., 2022).

Recently, many efforts have been made to apply transformers to handle graph data in the transportation domain, including traffic speed, flow prediction, and traffic demand prediction (Guo et al., 2022; Wen et al., 2023; Yan et al., 2022; Yang et al., 2023). While transformers have been used for time series imputation (Du et al., 2023; Zhang et al., 2024), their direct application to our spatiotemporal kriging problem remains challenging. Unlike missing value imputation, which relies partly on historical data for node knowledge, kriging interpolation necessitates estimating data for nodes devoid of any historical records—a more intricate task demanding a holistic consideration of spatiotemporal relationships. Notably, there's limited research applying transformers specifically to address the kriging interpolation problem. The Kriformer introduces a global attention mechanism designed to handle intricate networks encompassing myriad sensor nodes. This mechanism enhances the model's expressive capacity, overcoming limitations associated with local message passing in conventional methods. Consequently, it mitigates over-smoothing issues, preserving the original node features more effectively.

## 3. Problem Setup
### 3.1 Spatiotemporal Kriging Problem

Consider a sensor network represented by a directed graph $G = (V, E, A)$, where $V \equiv \{v_1, v_2, \dots, v_N\}$ represents the $N$ nodes in the system where sensors are installed and monitored, $E$ represents the connection relationships between adjacent nodes, such that $e_{ij} = (v_i, v_j)$ denotes an edge or an anticipated edge from node $v_i$ to node $v_j$, and $A$ denotes an adjacency matrix used to represent the strength of association between nodes, such as a matrix of distances.

In practice, sensor owners or managers may face scenarios where sensors cannot cover the entire network range due to high installation costs or may become dysfunctional, resulting in increased maintenance expenses; in such situations, they seek to interpolate time-series sensor data on unmonitored or malfunctioning nodes using data from operational sensor locations.[1] Note that in constructing data on unmonitored sensors, we need to establish the spatial correlation edge connections between unobserved nodes and observed nodes and capture the temporal relationship. This practical demand formulates the spatiotemporal kriging problem (see Figure 2).

To formally describe our spatiotemporal kriging problem, first define a subset $V_o \equiv \{v_1, \dots, v_O\} \subset V$ containing $O$ nodes in which node has observed sensor data and a subset $V_u \equiv \{v_1, \dots, v_U\} \subset V$ containing $U$ nodes in which data is totally unobserved. Figure 2 visually demonstrates the reconstruction process of the kriging problem. Within any time period $(t, t+T]$ of length $T$, the nodes in $V_o$ (purple nodes in the figure) are equipped with sensors, containing sensor data at time slots $t+1, t+2, \dots, t+T$. These data are denoted as a tensor $X_{t,T}^{V_o} \equiv \{X_{t+s}^{V_o} \equiv \{x_{t+s}^1, \dots, x_{t+s}^k, \dots, x_{t+s}^N\}, s = 0, \dots, T-1\}$, where $x_{t+s}^k \in \mathbb{R}^C$ is the observed sensor data of node $k \in V_o$ at time $t+s$ with channel size $C$. Note that $X_{t,T}^{V_o}$ is of shape $\mathbb{R}^{T \times O \times C}$. The unobserved nodes in $V_u$ (depicted in grey in the figure) lack any available data. We denote the observations of all the nodes in $V_u$ within the $T$ time periods as a tensor $X_{t,T}^{V_u} \equiv \{X_{t+s}^{V_u} \equiv \{x_{t+s}^1, \dots, x_{t+s}^k, \dots, x_{t+s}^N\}, s = 0, \dots, T-1\}$, where $X_{t,T}^{V_u}$ is of shape $\mathbb{R}^{T \times U \times C}$. We set $X_{t,T}^{V_u} = \mathbf{0}$ for all the elements in the tensor.

The task of kriging is to utilize the above $X_{t,T}^{V_o}$ to generate complete data for all unobserved nodes, represented as $\tilde{X}_{t,T}^{V_u} \equiv \{\tilde{X}_{t+s}^{V_u} \equiv \{\tilde{x}_{t+s}^1, \ldots, \tilde{x}_{t+s}^k, \ldots, \tilde{x}_{t+s}^N\}, s = 0, \ldots, T-1\}$, where $\tilde{x}_{t+s}^k \in R^C$ is the interpolated sensor data of node $k \in V_u$ at time $t+s$.

---

[1] Throughout this study, we interchangeably refer to these positions as "nodes," "sensors," or "locations" to accurately depict the problem's nature.)

### 3.2 Casting Kriging Problem as a supervised learning task

Inferring time-series data for unobserved nodes is a difficult task since all the ground-truth historical observations are missing. To tackle this, we propose utilizing measurements from other detected sensors for "spatial prediction". In other words, the measurements from spatially-connected nodes can serve as features or predictors to interpolate unobserved quantities. However, to make wise interpolation, the process still requires a label to guide the learning.

We introduce a random masking scheme to transform the kriging problem into a supervised learning task. The transformation process involves the following steps:

(1) Randomly select and mask some observed nodes in $V_o$ to create a set of masked observed nodes $V_{mo} \subset V_o$. The remaining unmasked observed nodes form a set of $V_{umo} \subset V_o$ and $V_{mo} \cup V_{umo} = V_o$.

(2) Set the masked observations in $V_{mo}$, denoted as $\bar{X}_{t,T}^{V_{mo}}$, to 0, while leaving the observations of unmasked nodes, $X_{t,T}^{V_{umo}}$, untouched.

(3) Fed $X_{t,T}^{V_{umo}}$, $\bar{X}_{t,T}^{V_{mo}} = \mathbf{0}$ and $X_{t,T}^{V_u} = \mathbf{0}$ into our proposed Kriformer model $f(\cdot)$ as predictor. The model generates the predicted values $\tilde{X}_{t,T}^{V_{umo}}$, $\tilde{X}_{t,T}^{V_{mo}}$ and $\tilde{X}_{t,T}^{V_u}$:

$$\tilde{X}_{t,T}^{V_{umo}}, \tilde{X}_{t,T}^{V_{mo}}, \tilde{X}_{t,T}^{V_u} = f(X_{t,T}^{V_{umo}}, \bar{X}_{t,T}^{V_{mo}} = \mathbf{0}, X_{t,T}^{V_u} = \mathbf{0}) \quad (1)$$

(4) Construct a reconstruction loss function by comparing the predicted values with the ground-truth values using Mean Squared Error (MSE):

$$\mathcal{L} = \left\| X_{t,T}^{V_{umo}} - \tilde{X}_{t,T}^{V_{umo}} \right\|_2^2 + \left\| X_{t,T}^{V_{mo}} - \tilde{X}_{t,T}^{V_{mo}} \right\|_2^2 \quad (2)$$

It's crucial to note that the observations of masked nodes $V_{mo}$, $X_{t,T}^{V_{mo}}$, are available as labels in the loss function but are treated as unknown on the feature side.

Once the supervised learning task has been done, the trained model can be used to interpolate observations on unobserved nodes by $f(X_{t,T}^{V_{umo}}, X_{t,T}^{V_{mo}}, X_{t,T}^{V_u} = \mathbf{0})$.

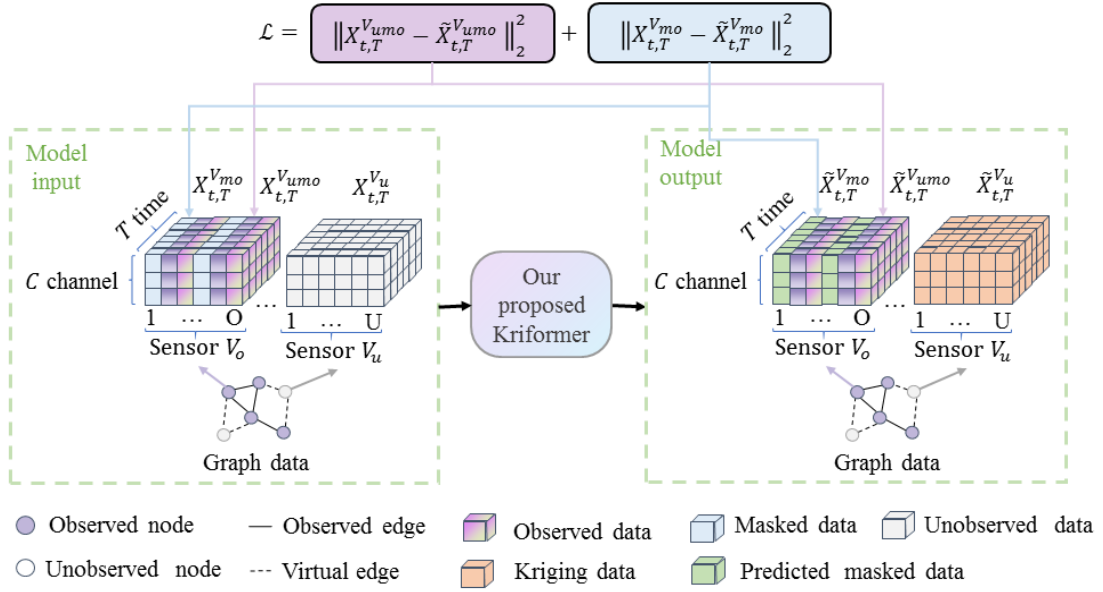

Figure 2: Illustration of the kriging process

**4. An Improved Transformer Model for Spatiotemporal Kriging Problem**

This section introduces our proposed Kriformer model, which is based on a transformer architecture tailored to address the kriging problem. Kriformer follows an encoder-decoder architecture, and its overall pipeline is depicted in Figure 3. We begin by describing the overall pipeline and the flow of training data within the different blocks of Kriformer in Section 4.1. The detailed operations of each block are further elaborated upon in Sections 4.2, 4.3, and 4.4.

**4.1 Overall Pipeline**

The overall pipeline includes the following four steps.

(1) **Data preprocessing stage.** At time $t$, the original encoder input is $X_{t,T}^V \in \mathbb{R}^{T \times N \times C}$ representing spatiotemporal data tensor containing observations from both observed and unobserved nodes for previous $T$ times. For the observed node set $V_o$, $X_{t,T}^{V_o} \in \mathbb{R}^{T \times O \times C}$ contains the observed sensor data; for the unobserved nodes, $X_{t,T}^{V_u} \in \mathbb{R}^{T \times U \times C}$ are set to be fully 0. The original input $X_{t,T}^V$ will first experience a masking scheme with an elementwise product $\bar{X}_{t,T}^V = X_{t,T}^V \odot M$ where $M$ comprises a random mask consisting of 0s and 1s.

(2) **Embedding process.** $\bar{X}_{t,T}^V$ will then undergoes a learnable linear projection to create input embedding $\hat{X}_{t,T}^V \in \mathbb{R}^{T \times N \times D}$, where $D$ is the embedding dimension. The input embedding $\hat{X}_{t,T}^V$ will then experience a Spatiotemporal Embedding (STE) block in which a spatiotemporal positional representation for diverse nodes across both temporal and spatial dimensions is elementwise added.

(3) **Encoding process.** The encoder contains $N_{en}$ sub-layers (encoder layers). Each sub-layer includes a Multi-head Temporal Attention (MTA) and a Multi-head Spatial Attention (MSA) targeting at capturing temporal and spatial correlations, respectively. Both are self-attention mechanisms (for a more detailed introduction to the attention mechanism, please refer to Appendix A). A residual skip connection and layer-wise normalization schemes are used in the same way as the original Transformer (Vaswani et al., 2017); see Figure 3. The representation outputted after the two attention mechanisms is fed into feed-forward network (FFN) to generate the output for the sub-layer. In sum, $\hat{X}^V_{t,T}$ along with the spatiotemporal positional embedding is forwarded to each sub-layer of the encoder, generating the encoder's output, $X^V_{t,T}(Encoder) \in \mathbb{R}^{T \times N \times D}$.

(4) **Decoding process.** The decoder contains $N_{de}$ sub-layers (decoder layers). The decoder receives two inputs: the decoder input $X \in \mathbb{R}^{T \times N \times C}$ and the encoder's output $X^V_{t,T}(Encoder) \in \mathbb{R}^{T \times N \times D}$. Similar to the encoder, the initial input $X \in \mathbb{R}^{T \times N \times C}$ is first through a learnable linear projection (output embedding) to generate $X' \in \mathbb{R}^{T \times N \times D}$ and then go to the two MTA and MSA self-attention layers. Moreover, the decoder also contains a Multi-head Spatial Interaction Attention layer (MSIA) where $X^V_{t,T}(Encoder)$ serves as keys and values and the $X'$ are the queries. The representation outputted after the two self-attention layers and the attention layer is fed into FFN to generate the output for the sub-layer. The output from the final sub-layer will be projected via a linear layer to produce expected predictions for virtual nodes.

Kriformer integrates three distinct multi-head attention modules: Multi-head Temporal Attention (MTA), Multi-head Spatial Attention (MSA), and Multi-head Spatial Interaction Attention (MSIA). We have redesigned the sub-layers (namely encoder/decoder layers) of the encoder and decoder within the transformer architecture. Specifically, in each sub-layer (illustrated in grey rectangles), the lower layers (depicted in orange rectangles) comprise stacked attention modules exclusively, while the upper layers (shown in blue rectangles) solely contain Feed-Forward Networks (FFN). This arrangement facilitates the swift propagation of states from the temporal attention module to the spatial attention module. To offer an insight into Kriformer's workflow, let's delve into the details of its data pipeline:

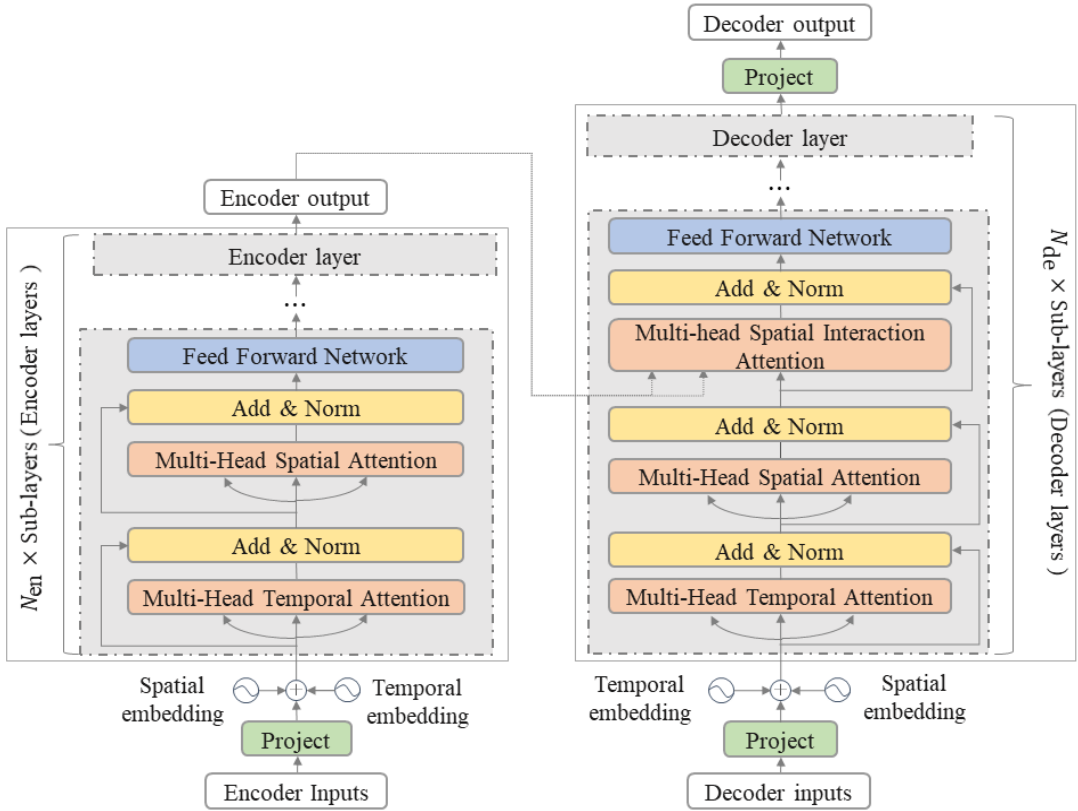

Figure 3: The overall model architecture of Kriformer, employing an encoder (left) - decoder (right) structure designed specifically to tackle the spatiotemporal kriging problem within traffic scenarios. The model harnesses Multi-Head Temporal Attention (MTA) and Multi-Head Spatial Attention

(MSA) mechanisms, specifically crafted to capture temporal and spatial autocorrelations. Additionally, it integrates Multi-head Spatial Interaction Attention (MSIA) to effectively capture spatial correlations between input and output. The Feed-Forward Network (FFN) comprises two layers utilized for forward propagation.

**4.2 Spatial–Temporal Embeddings (STE) block**

Graph Neural Networks (GNN) grapple with a restricted receptive field due to their localized message-passing mechanism, resulting in over-smoothing, particularly in deep GNN. In graph-structured data, relevant nodes might lie distantly from the central node, and nodes within the neighborhood may contain noise, thereby undermining the effectiveness of GNN. Transformers, adept at aggregating information from distant nodes and capturing long-term sequential dependencies, however, lack inherent positional information integration within their various attention models. To address this, we propose the Spatial–Temporal Embeddings (STE) module, as shown in Figure 4. This module aims to convert discrete symbols (e.g., temporal or spatial nodes) into continuous vector representations, enabling more effective utilization of graph structural information within Kriformer's sub-layers. These representations guide diverse attention modules for improved modeling of temporal and spatial features. In terms of the computation process, spatial and temporal information of nodes form spatial and temporal embeddings, respectively. These embeddings are coupled at the end of the embedding layer to create STE, providing an independent parameter space for the spatiotemporal dimensions and guiding the learning of spatial and temporal attention mechanisms.

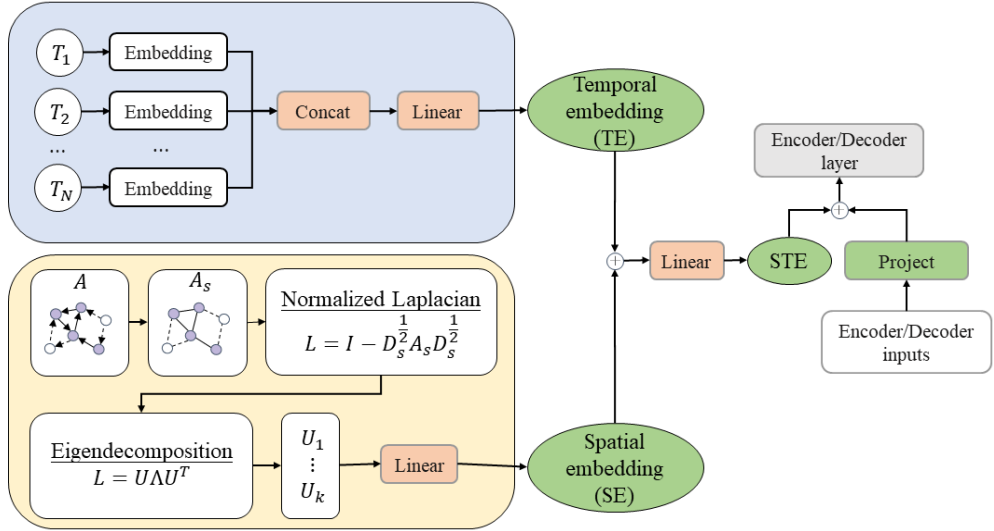

Figure 4: Spatiotemporal Embeddings (STE) diagram

4.2.1 Temporal Embeddings

Our temporal attention module operates solely through self-attention mechanisms to capture dynamics. However, in time series modeling, sequential information significantly impacts accuracy as nearby estimations often exhibit higher correlations. Explicitly incorporating sequence bias into the model can enhance estimations. To achieve this, each element in the input is furnished with a temporal embedding (TE) using sine and cosine functions, imparting similar positional features to neighboring elements. These temporal embeddings, though, neglect spatial effects, applying the

same temporal embedding operation to each spatial node, with the formula outlined below:

$$TE_{(t,2d)} = \sin(t/10000^{2d/D})$$
$$TE_{(t,2d+1)} = \cos(t/10000^{2d/D})$$
(3)

The index $t$ denotes the temporal position of each element within the time slots, while $d$ signifies the feature dimension index, totaling $D$ dimensions. To generate embeddings for different time positions $t$, temporal embedding employs sine functions for even embedding indices and cosine functions for odd embedding indices. Consequently, the temporal embedding yields an output represented as $TE \in \mathbb{R}^{T \times D}$.

### 4.2.2 Spatial Embeddings

We introduce Eigenmaps (Dwivedi and Bresson, 2020), a graph embedding technique that encodes structural information into spatial embeddings (SE). Eigenmaps leverage graph structures by precomputing Laplacian eigenvectors derived from available graph information to represent node positions. As these eigenvectors extend the concept of SE from the original transformer, better encapsulating distance-aware information—where nearby nodes share similar position features while distant ones exhibit dissimilarities—we adopt computed Laplacian eigenvectors as the SE in our Kriformer.

To compute spatial embeddings via Eigenmaps, we follow these steps:

(1) Symmetrize the adjacency matrix $A$, denoted as $A_s = \max(A, A^T)$, assuming the generated undirected graph remains connected.

(2) Compute the normalized Laplacian matrix $L = I - D_s^{-1/2} A D_s^{-1/2}$, where $D_s$ is the diagonal degree matrix of $A_s$, fulfilling $D_s(i,j) = \sum_{j=1}^{N} A_s(i,j)$, and $I$ represents the identity matrix.

(3) Perform an eigendecomposition $L = U \Lambda U^T$ on the Laplacian matrix to derive the matrix of eigenvectors U, where $U = (v_0, v_1, \ldots, v_{N-1})$, and $\Lambda = \text{diag}(\lambda_0, \lambda_1, \ldots, \lambda_{N-1})$ is the eigenvalue matrix, ensuring $0 < \lambda_0 \ll \cdots \ll \lambda_{N-1}$.

(4) Construct the k-dimensional embedding ($k < N$) for the node $v_i$ as $SE_i = [v_1(i), v_2(i), \ldots, v_k(i)] \in \mathbb{R}^k$.

Following the Eigenmaps computation, the final spatial embedding $SE \in \mathbb{R}^{N \times D}$ for all nodes is generated through a learnable linear transformation operation.

### 4.2.3 Merging spatial and temporal embeddings

To facilitate merging the Temporal Embedding (TE) and Spatial Embedding (SE), aligning their dimensional spaces becomes essential. Initially, $TE \in \mathbb{R}^{T \times d}$ undergoes spatial dimension expansion to $TE \in \mathbb{R}^{T \times N \times D}$ through $N$ operations. Correspondingly, $SE \in \mathbb{R}^{N \times D}$ transforms into $SE \in \mathbb{R}^{T \times N \times D}$ by temporal dimension replication $T$ times. Finally, TE and SE are coupled (e.g., concatenated, added, or multiplied) and linearly mapping them into $STE \in \mathbb{R}^{T \times N \times D}$.

## 4.3 Encoder

The encoder's function is to take in inputs and transform them into higher-level contextual representations, capturing the spatiotemporal information of the inputs. Within Kriformer, the encoder comprises an input projection layer and $N_{en}$ identical encoder layers with skip connections. Each encoder layer comprises three key components: (i) a Multi-Head Temporal Attention (MTA);

(ii) a Multi-Head Spatial Attention (MSA); and (iii) a feed-forward network (FFN). The MTA and MSA are multi-head structures designed for capturing temporal and spatial correlations, respectively, while the FFN manages positional transformations. These interconnected components collaboratively learn spatiotemporal representations in a sequential manner.

4.3.1 Multi-head Temporal Self-attention (MTA)

Within MTA, when the query, key, and value align as the same vector sequence, as elaborated in Appendix A, the multi-head attention transitions into multi-head self-attention, signifying $Q = K = V$. Consequently, along the temporal dimension, multi-head self-attention directly derives the $(l)$th layer $H^l \in \mathbb{R}^{T \times N \times D}$ from the output of the $(l-1)$th layer $H^{l-1} \in \mathbb{R}^{T \times N \times D}$. Different from natural language processing, which handles variable-length inputs, our inputs maintain a fixed length. Hence, we omit padding masks, as the encoder's sole responsibility lies in encoding the input sequence, excluding involvement in generation or decoding processes. Additionally, each attention mechanism head possesses its distinct weight matrix $W$. Algorithm 1 illustrates the procedural details of MTA.

---

**Algorithm 1:** MTA for encoder.

**Input:** $H^{l-1} \in \mathbb{R}^{T \times N \times D}$
**Output:** $H^l \in \mathbb{R}^{T \times N \times D}$

for $i$ in $1:N$ do
  for $h$ in $1:N_h$ do
    $W_{i,h}^q \leftarrow H_{i,h}^{l-1}$;
    $W_{i,h}^k \leftarrow H_{i,h}^{l-1}$;
    $W_{i,h}^v \leftarrow H_{i,h}^{l-1}$;
    $Q_{i,h} \leftarrow W_{i,h}^q H_{i,h}^{l-1}$;
    $K_{i,h} \leftarrow W_{i,h}^k H_{i,h}^{l-1}$;
    $V_{i,h} \leftarrow W_{i,h}^v H_{i,h}^{l-1}$;
    $Head_{i,h} \leftarrow softmax(\frac{Q_{i,h} K_{i,h}^T}{\sqrt{D/N_h}}) V_{i,h}$;
  end
  $H_{i,:}^l \leftarrow Concat(Head_{i,h}, \ldots, Head_{i,N_h}) W_O$;
end
$H^l \leftarrow LayerNorm(H^{l-1} + H^l)$;

**Result:** $H^l$

---

4.3.2 Multi-head Spatial Self-attention (MSA)

The operation of MSA differs from MTA. In MSA, spatial node masking is introduced, similar to the spatial node masking technique in Graph Attention Network (Veličković et al., 2017). When the correlation between two nodes is extremely weak (with a strength of 0), the corresponding position will be replaced by a maximum negative value (-∞), making the attention score for that position tend toward 0. This process effectively eliminates interference from irrelevant nodes during the learning of spatial correlations via the SoftMax function. Subsequently, these matrices are concatenated and linearly projected using scale dot-product attention. Finally, at the end of MSA, skip connections and layer normalization techniques are employed. For the detailed execution of

MSA, refer to Algorithm 2.

---

**Algorithm 2:** MSA for encoder.

**Input:** $H^{l-1} \in \mathbb{R}^{T \times N \times D}$
**Output:** $H^l \in \mathbb{R}^{T \times N \times D}$

---

**for** $t$ in $1:T$ **do**
  **for** $h$ in $1:N_h$ **do**
    $W_{t,h}^q \leftarrow H_{t,h}^{l-1}$;
    $W_{t,h}^k \leftarrow H_{t,h}^{l-1}$;
    $W_{t,h}^v \leftarrow H_{t,h}^{l-1}$;
    $Q_{t,h} \leftarrow W_{t,h}^q H_{t,h}^{l-1}$;
    $K_{t,h} \leftarrow W_{t,h}^k H_{t,h}^{l-1}$;
    $V_{t,h} \leftarrow W_{t,h}^v H_{t,h}^{l-1}$;
    $Head_{t,h} \leftarrow softmax(\frac{Q_{t,h} K_{t,h}^T}{\sqrt{D/N_h}} + mask) V_{t,h}$;
  **end**
  $H_{t,:}^l \leftarrow Concat(Head_{t,h}, \ldots, Head_{t,N_h}) W_O$;
**end**
$H^l \leftarrow LayerNorm(H^{l-1} + H^l)$;

---

**Result:** $H^l$

### 4.3.3 Feed Forward Network (FFN)

The FFN functions as a forward propagation layer, comprising two densely connected layers employed as tensor dimension transform. Analogous to the MTA and MSA, the FFN integrates methodologies such as skip connections and layer normalization. Its specific structure is outlined as follows:

$$H^l = LayerNorm(H^{l-1} + ReLU(W_1 H^{l-1} + b_1) W_2 + b_2) \tag{4}$$

where $W_1, W_2 \in R^{D \times D}$, $b_1, b_2 \in \mathbb{R}^D$ represent the trainable parameters.

### 4.4 Decoder

The decoder's main role is to incrementally generate the targets, drawing upon both the generated content and the contextual representation furnished by the encoder. The decoder stack comprises an input projection layer, $N_{de}$ identical decoder layers with skip connections, and a linear layer for projection. Each decoder layer comprises four components: MTA, MSA, MSIA, and FFN. Within the decoder, we've developed two distinct types of spatial attention modules: namely, MSA and MSIA modules. The MSA in the decoder remains entirely identical to the one described in Algorithm 2, aimed at capturing spatial correlations.

### 4.4.1 Multi-head Spatial Interaction Attention (MSIA)

For enhanced capture of attentional distribution relationships between observed and unobserved nodes, the MSIA has been introduced. MSIA utilizes a tensor $X_{en} \in \mathbb{R}^{T \times N \times D}$ derived from the encoder's output as keys and values, while employing another tensor $X \in \mathbb{R}^{T \times N \times D}$ from the decoder's MTA as queries. MSIA serves as a bridge between the encoder's output and each decoder layer, allowing selective attention to spatial and temporal information of different nodes based on the currently generated content and the encoded information from the encoder. Unlike the MSA, MSIA implements separate queries, keys, and values, signifying a distinctive characteristic.

---

**Algorithm 3:** MSIA for decoder.

**Input:** $H^{l-1}(encoder) \in \mathbb{R}^{T \times N \times D}$, $H^{l-1}(decoder) \in \mathbb{R}^{T \times N \times D}$
**Output:** $H^l \in \mathbb{R}^{T \times N \times D}$

for $t$ in $1:T$ do
  for $h$ in $1:N_h$ do
    $W_{t,h}^q \leftarrow H_{t,h}^{l-1}(decoder);$
    $W_{t,h}^k \leftarrow H_{t,h}^{l-1}(encoder);$
    $W_{t,h}^v \leftarrow H_{t,h}^{l-1}(encoder);$
    $Q_{t,h} \leftarrow W_{t,h}^q H_{t,h}^{l-1}(decoder);$
    $K_{t,h} \leftarrow W_{t,h}^k H_{t,h}^{l-1}(encoder);$
    $V_{t,h} \leftarrow W_{t,h}^v H_{t,h}^{l-1}(encoder);$
    $Head_{t,h} \leftarrow softmax(\frac{Q_{t,h} K_{t,h}^T}{\sqrt{D/N_h}}) V_{t,h};$
  end
  $H_{t,:}^l \leftarrow Concat(Head_{t,h}, \ldots, Head_{t,N_h}) W_O;$
end
$H^l \leftarrow LayerNorm(H^{l-1} + H^l);$

**Result:** $H^l$

---

## 5. Experiments

In this section, we conducted experiments on two real-world traffic speed datasets to evaluate the kriging performance of the proposed Kriformer model. Initially, we provided a brief description of the datasets, outlining the experimental settings and details such as advanced kriging baseline models. Subsequently, we compared the kriging performance across various models. Finally, we conducted several ablation studies to perform model analysis.

### 5.1 Dataset and Preprocessing

5.1.1 Dataset

Extensive experiments were conducted on two public datasets: PEMS-LA and PEMS-Bay, both sourced from the Performance Measurement System (PEMS) by the California Transportation Agencies (CalTrans). The Caltrans PEMS system comprises over 15,000 detectors that collect real-time traffic data every 30 seconds. The original traffic records were aggregated at 5-minute intervals. Additionally, geographical location information about sensor stations was included in the dataset to compute distances between nodes. A summary of these datasets is provided in Table 1.

    (1) LA: The PEMS-LA dataset comprises traffic speed data gathered from 207 loop detectors in Los Angeles spanning four months (from Mar 1, 2012, to Jun 30, 2012) at 5-minute intervals.

    (2) Bay: The PEMS-Bay dataset contains traffic speed data collected over 6 months (from Jan 1, 2017, to May 13, 2017) at 5-minute intervals by 325 loop detectors in the San Francisco Bay Area.

    The input data underwent standardization, scaling to a unit interval by subtracting the mean

and dividing by the standard deviation, as shown below:

$$X = \frac{X - Mean(X)}{Std(X)} \tag{5}$$

where $Mean(X)$ and $Std(X)$ are the mean and standard deviation of the input data $X$, respectively.

Table 1: Data description table

| Dataset | Internal | Timestep | Nodes | Edges | Time span | Data missing |
|---|---|---|---|---|---|---|
| PEMS-LA | 5min | 34272 | 207 | 1515 | 4 month | 8.109% |
| PEMS-Bay | 5min | 52116 | 325 | 2691 | 6 month | 0.003% |

5.1.2 Adjacency Matrix

Based on the preceding description, the traffic network is represented by $G = (V, E, A)$, where the adjacency matrix $A \in \mathbb{R}^{N \times N}$ signifies the pairwise relationships among nodes in this network. The values within this matrix reflect the respective edge strengths. The strength between nodes is determined by distance information. However, directly using road distance to describe strength is unreasonable, despite the intuitive understanding that the strength between nodes tends to weaken as the distance between them increases. The accessibility between nodes is determined by the inherent structure of the traffic road network; therefore, we consider it as a representation of the spatial relationships between nodes in the adjacency matrix $A$. The formulation for the distance-based road network accessibility weight, denoted as $\omega$, is defined as follows:

$$\omega_{ij} = \begin{cases} \exp\left(-\frac{d_{ij}^2}{\sigma^2}\right) & i \neq j \text{ and } \frac{d_{ij}^2}{\sigma^2} > \varepsilon \\ 0 & otherwise \end{cases} \tag{6}$$

Where $d_{ij}$ represents the road network distance from the node $v_i$ to node $v_j$, The parameter $\sigma$ denotes the distance standard deviation, while $\varepsilon$ serves as the hyperparameter controlling the sparsity of the weighted matrix. This value is set to 0.1 in this paper.

**5.2 Experimental Details**

5.2.1 Hyperparameters Settings.

Hyperparameters refer to predefined parameters that remain fixed during the training process but have a significant impact on the final model results. In our paper, both the encoder and decoder have 2 layers of spatiotemporal attention, with 4 attention heads and a feature dimension $D$ of 64. A dropout layer is integrated into the model as a regularization technique to mitigate overfitting, with a dropout probability of 0.2 for randomly dropping neurons. The Adam optimization method is utilized for gradient updates during the training process, with a batch size of 8 and a learning rate of 0.001.

5.2.2 Baselines

In demonstrating the superiority of our proposed Kriformer model, we conducted a comprehensive comparison against several state-of-the-art baseline methodologies. These encompass traditional kriging models, tensor factorization methods, and advanced deep learning approaches, as presented in Table 2.

(5) Traditional methods:
- KNN (K Nearest Neighbor): Extends readings by averaging values from the k nearest spatial neighbors of the missing station (Hudak et al., 2008).
- Okriging: A geo-statistical interpolation method assuming spatial correlation based on distances between points, elucidating surface variation (Stein, 1999), which is provided by a Python framework (downloaded from https://github.com/GeoStat-Framework/PyKrige).

(2) Tensor factorization methods:
- BATF (Bayesian Augmented Tensor Factorization): A comprehensive Bayesian framework proposed for imputing missing traffic speed data (Chen et al., 2019).
- LETC (Laplacian Enhanced low-rank Tensor Completion): Considers spatiotemporal correlations in large-scale networks using low-rank tensor methodologies for spatiotemporal kriging (Nie et al., 2023).

(5) Deep learning methods:
- GCN-GRU: A modified variant of Kriformer, replacing the graph transformer with GCN (Graph Convolutional Network) and GRU (Gated Recurrent Unit).
- IGNNK (Inductive Graph Neural Network Kriging): Employs stacked diffusion graph convolution layers and implements a random mask training strategy for spatial kriging tasks (Wu et al., 2021).

Table 2: Model classification

| Model | Types | Spatial | Temporal | Global attention | Long-range correlations |
|---|---|---|---|---|---|
| KNN | Traditional methods | √ | × | × | × |
| Okriging | Traditional methods | √ | × | × | × |
| BATF | Tensor factorization methods | √ | √ | × | × |
| LETC | Tensor factorization methods | √ | √ | × | × |
| GCN-GRU | Deep learning methods | √ | √ | × | × |
| IGNNK | Deep learning methods | √ | √ | × | × |
| Kriformer | Deep learning methods | √ | √ | √ | √ |

5.2.3 Evaluation Metrics

During each iteration of model training, the mask rate of observed nodes for the random masking strategy is set at 30%. To gauge the model's performance, we conducted tests across various observation conditions: randomly masking 30%, 50%, and 70% of ground truth values to simulate unobserved locations (designated as SM3, SM5, SM7, respectively), and designating 30% of time intervals as unobserved periods during testing. Performance evaluation involved comparing estimated values against the masked ones. To assess our model's performance and benchmark it against other methods, we employed the following three metrics: MAE (Mean Absolute Error),

RMSE (Root Mean Squared Error), and MAPE (Median Absolute Percentage Error).

$$MAE = \frac{1}{NP}\sum_{i=0}^{N}\sum_{t=1}^{P}|x_{i,t} - \hat{x}_{i,t}| \tag{7}$$

$$RMSE = \sqrt{\frac{1}{NP}\sum_{i=0}^{N}\sum_{t=1}^{P}(x_{i,t} - \hat{x}_{i,t})^2} \tag{8}$$

$$MAPE = \frac{1}{NP}\sum_{i=0}^{N}\sum_{t=1}^{P}\left|\frac{x_{i,t} - \hat{x}_{i,t}}{x_{i,t}}\right| \times 100\% \tag{9}$$

where $\hat{x}_{i,t}$ represents the estimated value and $x_{i,t}$ denotes the ground truth for location $i$ at time step $t$.

### 5.3 Experimental Results
#### 5.3.1 Model Performance Comparison

Tables 3 and 4 present a comparative analysis of kriging performance between Kriformer and various baseline models. Notably, Kriformer consistently demonstrates superior kriging results across all test scenarios. Most models experience deteriorating performance as the number of test nodes varies, yet Kriformer stands out for its robustness. This emphasizes the significance of concurrently modeling global and local correlations. The incorporation of global information beyond adjacent nodes furnishes more referenceable features for the target node and enables Kriformer to achieve high-precision kriging. Traditional statistical methods tend to overlook dynamic spatiotemporal information in traffic networks, leading to inaccuracies in traffic data estimation. BATF excels in imputing missing traffic speed data through tensor factorization but exhibits moderate performance in kriging tasks. LETC, designed to address kriging issues by considering global spatial and temporal correlations, struggles with datasets smaller than their intended spatial dimensions. In contrast, within deep learning approaches, IGNNK outperforms statistical methods and tensor decomposition in kriging tasks by leveraging crucial information through a graph neural network's random walk mechanism, despite focusing solely on local spatial correlations. Conversely, GCN-GRU's utilization of global graph data leads to message propagation from noisy, unrelated nodes, resulting in larger estimation errors due to learned representations containing erroneous information. In summary, effective aggregation of spatiotemporal information, guided by spatiotemporal attention mechanisms and position embeddings, significantly enhances kriging performance by leveraging pertinent global network data.

Table 3 showcases the kriging performance of different models under various settings of the LA dataset. Bold indicates the best results.

|  | LA (SM3) | | | LA (SM5) | | | LA (SM7) | | |
|---|---|---|---|---|---|---|---|---|---|
| Model | MAE | RMSE | MAPE(%) | MAE | RMSE | MAPE(%) | MAE | RMSE | MAPE(%) |
| KNN | 9.3451 | 12.6183 | 16.0636 | 9.0645 | 12.3969 | 15.5932 | 9.2095 | 12.3268 | 15.8257 |
| Okriging | 8.8305 | 12.5189 | 15.1791 | 8.6057 | 12.3125 | 14.8041 | 8.5778 | 12.2114 | 14.7401 |
| BATF | 8.9224 | 12.5163 | 15.3369 | 8.6973 | 12.3296 | 14.9616 | 8.6279 | 12.1513 | 14.8261 |
| LETC | 7.7938 | 10.9501 | 13.3970 | 7.8746 | 11.2145 | 13.5464 | 8.2690 | 11.5952 | 14.2095 |
| GCN-GRU | 7.6350 | 12.4798 | 13.1242 | 7.7435 | 12.3173 | 13.3209 | 7.8055 | 12.4648 | 13.4132 |
| IGNNK | 6.7511 | 10.3863 | 11.6047 | 7.2046 | 10.8682 | 12.3939 | 7.9161 | 11.5365 | 13.6031 |
| Kriformer | **5.8754** | **9.1950** | **10.0995** | **6.4159** | **10.1346** | **11.0370** | **6.9557** | **10.7374** | **11.9528** |

Table 4 showcases the kriging performance of different models under various settings of the BAY dataset. Bold indicates the best results.

|  | BAY (SM3) | | | BAY (SM5) | | | BAY (SM7) | | |
|---|---|---|---|---|---|---|---|---|---|
| Model | MAE | RMSE | MAPE(%) | MAE | RMSE | MAPE(%) | MAE | RMSE | MAPE(%) |
| KNN | 4.8852 | 8.1077 | 7.8171 | 4.9550 | 8.1061 | 7.9212 | 4.9836 | 8.1915 | 7.9812 |
| Okriging | 4.8749 | 8.0885 | 7.8006 | 4.9512 | 8.0933 | 7.9152 | 4.9850 | 8.1757 | 7.9833 |
| BATF | 4.8743 | 8.0555 | 7.7940 | 4.9740 | 8.1336 | 7.9460 | 4.9933 | 8.1841 | 7.9911 |
| LETC | 4.3481 | 7.2324 | 6.9526 | 4.4059 | 7.2571 | 7.0384 | 4.5992 | 7.6029 | 7.3603 |
| GCN-GRU | 4.5494 | 8.2090 | 7.2797 | 4.3960 | 7.9089 | 7.0276 | 4.4066 | 7.9204 | 7.0571 |
| IGNNK | 3.9285 | 6.8362 | 6.2862 | 3.9581 | 7.0519 | 6.3276 | 4.0535 | 7.3749 | 6.4915 |
| Kriformer | **3.6708** | **6.2510** | **5.8738** | **3.8070** | **6.7093** | **6.0859** | **3.9315** | **7.0398** | **6.2962** |

### 5.3.2 Temporal Visualization of Kriging Results

Figures 5 illustrate time-series results for the kriging problem with various settings on the LA and BAY datasets, presenting a visual representation of speed estimation performance at unobserved nodes. Each figure depicts the kriging estimated values (illustrated in orange) and the ground truth values (shown in blue) for detectors randomly chosen from the set of unmeasured locations across different time intervals, with the proportion of unobserved data ranging from 30% (SM3) to 70% (SM7). Overall, Kriformer demonstrates precise time-series estimation for unobserved locations, even under low observation rates (e.g., see the kriging results for SM7 in the third row, where the orange line closely follows the blue line). Analyzing kriging outcomes in the LA dataset reveals substantial fluctuations in the time series, with numerous traffic speed values abruptly dropping to zero. This occurs due to the inherently higher percentage of missing values in the LA dataset, reaching up to 8.109%. This emphasizes that our proposed Kriformer model maintains robust kriging performance despite the prevalence of substantial missing data, attributed to its consideration of global network information. In contrast, the BAY dataset displays lower instances of missing values, resulting in superior estimation performance despite the existence of data gaps.

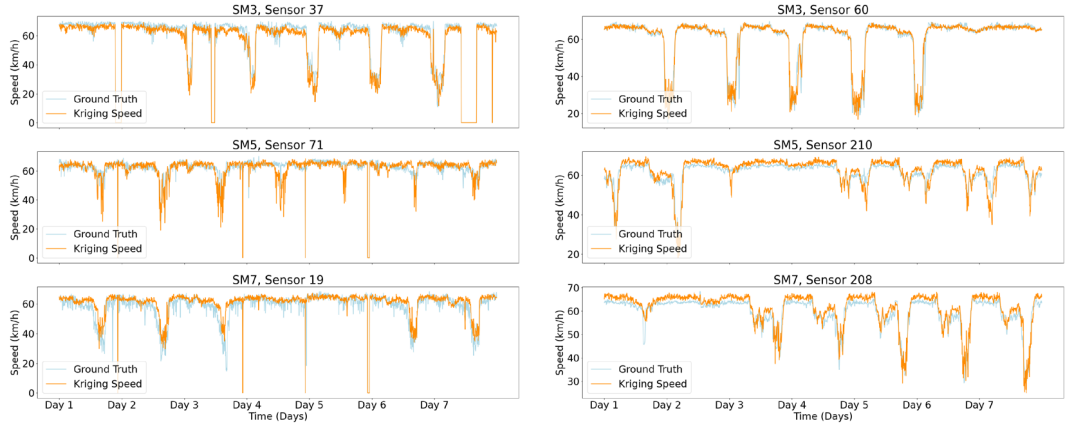

(a) Kriging results for the LA dataset  (b) Kriging results for the BAY dataset

Figure 5: Visualization of kriging results for the LA and BAY dataset. A comparison between actual and estimated values is presented for three randomly selected sensors with different temporal patterns.

A more nuanced comparison between Kriformer and two other state-of-the-art models concerning their time series results is presented in Figure 6. The findings indicate that during periods of relatively stable traffic speed, all three models provide estimations closely aligning with the actual values. However, upon closer scrutiny, LETC struggles to effectively capture fluctuations in traffic speed. This phenomenon may stem from LETC's need for refined spatiotemporal feature selection, where capturing complete global spatiotemporal information might blend in noise from some nodes, thereby diminishing the unique spatiotemporal features of the target node and resulting in kriging estimates that lean towards overall trends. In contrast, IGNNK focuses solely on local neighboring information and adeptly captures speed changes in nodes. Yet, it was observed that IGNNK's estimates exhibit considerable fluctuation errors when minor trends in node speed occur. This heightened sensitivity to variations in nearby nodes leads to a significant impact on estimations for the target node, even from minor changes in specific nodes.

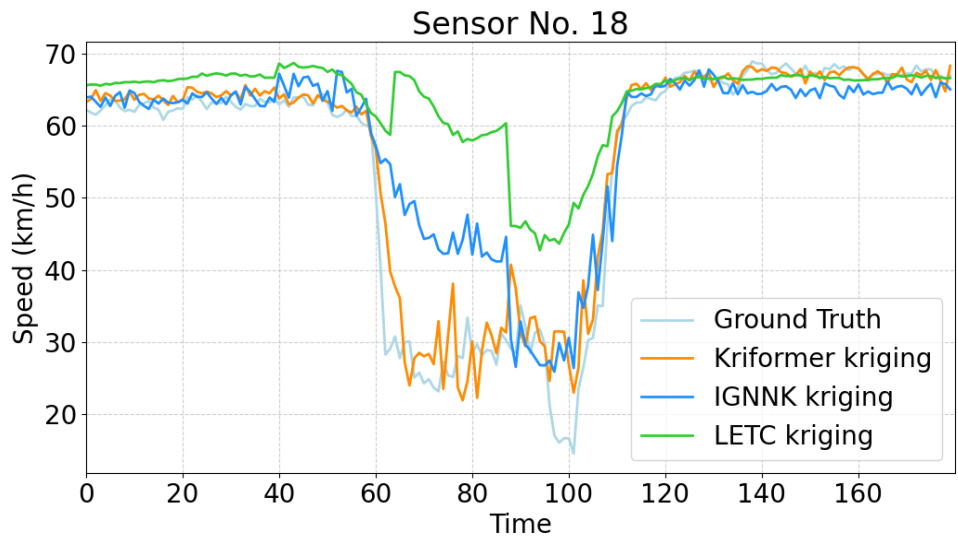

Figure 6: Detailed comparison of different kriging methods

### 5.3.3 Spatial Visualization of Kriging Results

Figure 7 portrays the ground truth speed values within the Los Angeles road network, alongside the kriging speed values generated by Kriformer and the estimation residuals (depicting the absolute differences between actual and estimated values), highlighting the efficacy of our proposed model. It's evident that Kriformer not only captures the global spatial distribution of traffic speeds but also discerns changes induced by local congestion. Comparing Figures 7(a), and (b), the superiority of the Kriformer model becomes apparent, especially in reflecting speed changes within locally congested areas. Figure 7(c) demonstrates a commendable alignment between Kriformer estimates and actual speed values, with over 90% of residual values falling below 5. Despite Kriformer's overall proficient estimation performance, errors are more pronounced in regions with few or no adjacent sensors (e.g., the single node in the lower-left corner of Figure 7(c)) or in highly congested areas with drastic traffic speed fluctuations. Despite considering spatiotemporal correlations among global nodes, information aggregation for target nodes primarily relies on neighboring sensors. Therefore, estimation becomes more challenging in areas with limited reference information. Overall, the sensor-level errors produced by Kriformer tend to be smaller than those of baseline models, particularly for sensors situated in congested areas.

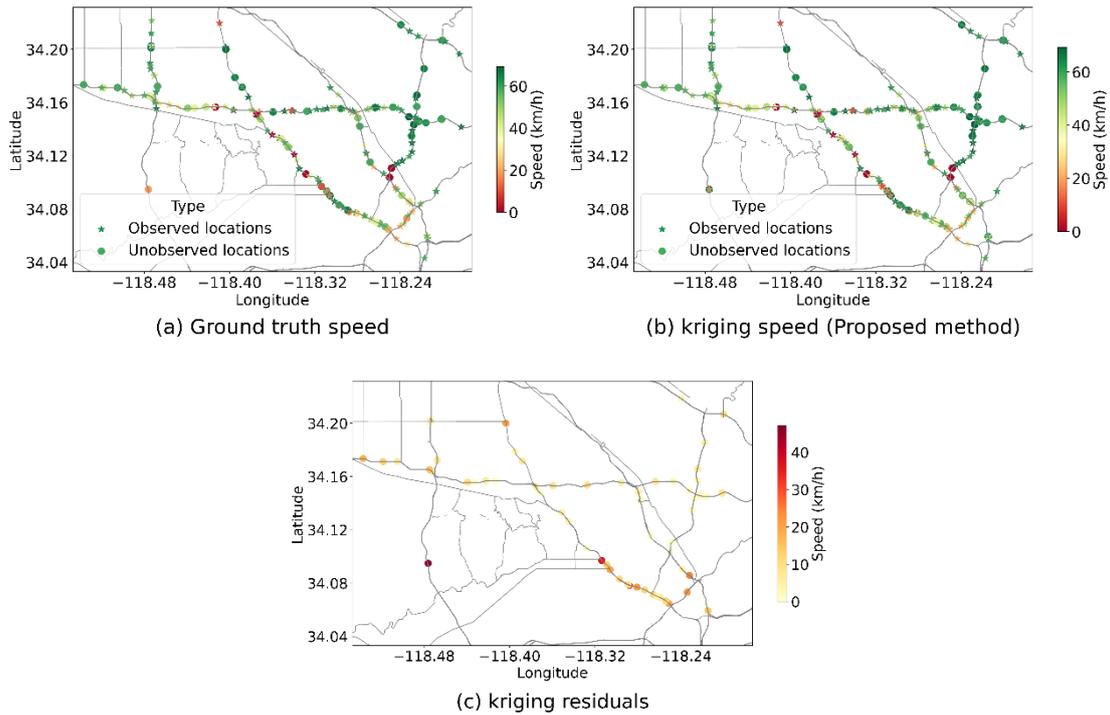

Figure 7: Kriging results of Kriformer on the Los Angeles road network. (a) Ground truth values. (b) Estimated traffic speeds from Kriformer. (c) Kriging residuals of Kriformer for each sensor.

### 5.4 Ablation Study

In this section, ablation experiments were conducted to assess the significance of each component within the proposed model, Kriformer, by comparing the contributions of various modules. Each module's impact was evaluated by creating new variant models of Kriformer wherein the corresponding module was removed. These variants were named as follows:

No TE: Spatial-Temporal Embedding (STE) contains only spatial embeddings, excluding sine-cosine functions, allowing assessment of the importance of temporal embeddings (TE).

No SE: STE includes only temporal embeddings without utilizing Eigenmaps technology, evaluating the significance of spatial embeddings (SE).

No STE: Removal of all STE components from Kriformer, leading to the model devolving into a multi-head attention mechanism focusing on spatiotemporal correlations.

No MTA: Multi-head temporal self-attention (MTA) is not employed within the encoder and decoder for modeling spatiotemporal correlations, assessing the importance of considering temporal correlations.

No MSA: Multi-head spatial self-attention (MSA) is omitted within the encoder and decoder for modeling spatiotemporal correlations, evaluating the importance of considering spatial correlations.

No MSIA: Multi-head Spatial Interaction Attention (MSIA) is excluded within the encoder and decoder for modeling spatiotemporal correlations, assessing the importance of considering spatial interaction correlations.

The performance evaluation of these variant models based on Kriformer was conducted on the LA dataset, as illustrated in Figure 8. Subsequently, we provide detailed insights into the significance of spatiotemporal embeddings and correlations.

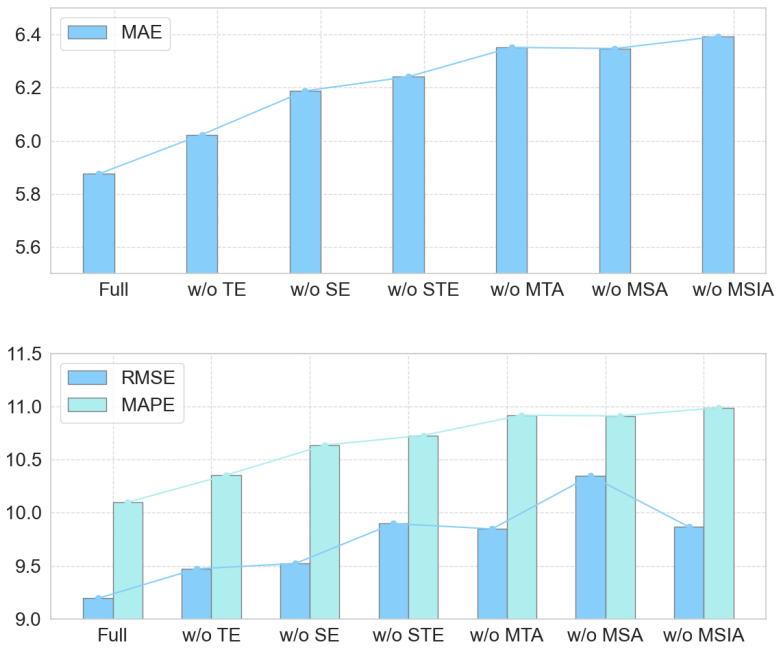

Figure 8: Ablation experiment results on LA dataset

From the perspective of spatiotemporal embeddings, Kriformer utilizes sine-cosine functions and Eigenmaps for temporal and spatial embeddings, coupled into the STE block. The absence of TE and SE modules results in a notable decrease in estimation performance. On the LA dataset, removing TE leads to reductions of 2.51%, 3.02%, and 2.51% in MAE, RMSE, and MAPE, respectively. Similarly, excluding SE showcases declines of 5.32%, 3.55%, and 5.32%, while the absence of STE indicates reductions of 6.22%, 7.64%, and 6.22%, respectively. These outcomes demonstrate the effectiveness of constructing spatiotemporal embedding modules, enhancing the kriging interpolation of traffic speed data. Particularly, the spatial embedding module exhibits a more pronounced enhancement in traffic kriging performance compared to the temporal embedding module. The spatial graph Laplacian forms the cornerstone of spatiotemporal embeddings within our Kriformer model, directly derived from network topology. As network scales increase, the adjacency matrix encompasses more intricate spatial relationships, reducing the impact of local anomalies. Therefore, for comprehensive consideration of global information, employing an effective graph embedding technique becomes crucial to guide the model's learning, offering more informative neighborhoods for unobserved nodes to reference observed values.

In the realm of spatiotemporal correlations, the transformer architecture's essence resides in the multi-head attention mechanism, renowned for its proficiency in interpreting multiple distribution patterns. Our proposed Kriformer model, an extension of the original transformer, incorporates three distinct attention modules: Multi-head temporal self-attention (MTA, see Algorithm 1), Multi-head spatial self-attention (MSA, see Algorithm 2), and Multi-head Spatial Interaction Attention (MSIA, see Algorithm 3). The absence of MTA resulted in declines of 8.10%, 7.10%, and 8.10% in MAE, RMSE, and MAPE on the LA dataset, respectively. MTA primarily focuses on capturing temporal dependencies, aiming to estimate the traffic speed of unobserved nodes by analyzing various time patterns in neighboring time series data from observed nodes. The performance degradation without MSA is similar to that without MTA, exhibiting degradation rates of 8.03%, 12.53%, and 8.03% on the LA dataset. Spatial attention predominantly deals with modeling in the spatial dimension, akin to time attention but tailored to comprehend spatial correlations within the graph, leveraging spatial knowledge. Furthermore, similar to the graph attention network, the spatial attention computation employs a masking process to filter weaker pairwise relationships in space, crucial for retaining node pairs with higher attention scores essential for kriging estimation concerning the target node.

The absence of MSIA resulted in declines of 8.80%, 7.30%, and 8.80%. MSIA differs from MSA in that it receives tensors from the encoder's output, used separately as keys and values. These keys gauge the correlation between the current decoder position and various positions within the input, while the values offer a comprehensive representation of the input, accessible to the decoder for reference and utilization during generation. Concurrently, tensors produced in the decoder's MTA serve as queries or content for the current position. MSIA acts as a crucial link between the encoder and decoder layers, facilitating selective attention to diverse spatial and temporal features within the input information, leveraging both the ongoing generated content and encoded details from the encoder. This enhancement significantly bolsters the model's performance and efficacy.

**5.5 Impact of Random Masking Ratio**

In our proposed training scheme, we employed a random masking strategy. During each iteration, a portion of observed nodes is randomly masked as $V_{mo}$, and their data is set to zero, i.e., $\bar{X}_{t,T}^{V_{mo}} = \mathbf{0}$.

To evaluate the parameter sensitivity of the random masking strategy, we varied the percentage of masked nodes and recorded the errors in Kriging results, as shown in Figure 9. From these results, it can be observed that the error remains relatively stable when the masking ratio is below 60%, with the minimum error occurring at a 30% ratio. However, errors in Kriging increase significantly when the masking ratio exceeds 70%.

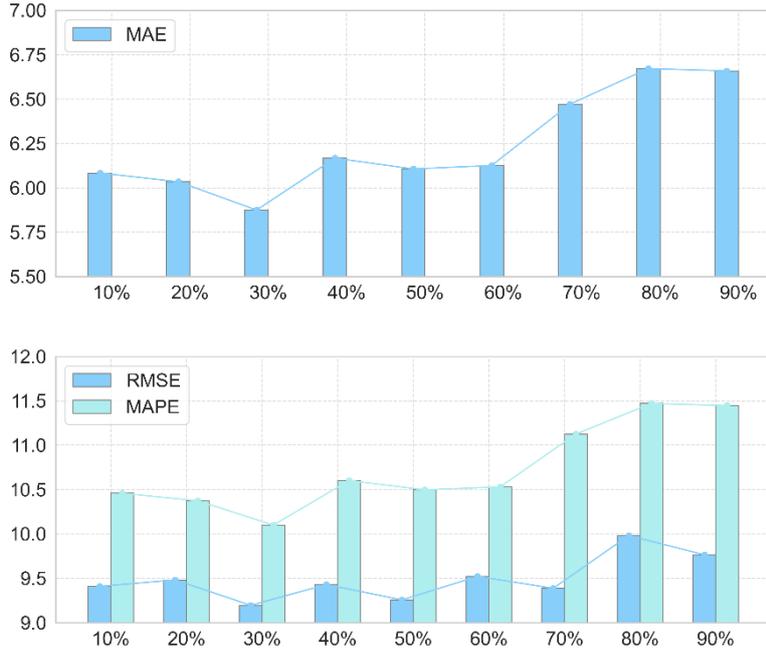

Figure 9: Impact of random masking ratio on LA dataset

## 6. Conclusion and Future Directions

Obtaining fine-grained spatiotemporal data for effectively understanding system dynamics is challenging due to sparse sensor coverage and data corruption. To address this, we introduce Kriformer, a kriging-based model designed for interpolating network data at locations lacking sensors under limited observations. Kriformer integrates a spatiotemporal embedding module that forms a universal representation of spatiotemporal nodes within the embedding layers. Meticulously designed attention layers in Kriformer cater to modeling both temporal and spatial aspects. Through experiments on two publicly available traffic speed datasets, we derive the following insights:

(1) Global spatial consideration is paramount: Kriformer's ability to consider spatiotemporal correlations at a global scale allows for learning robust patterns, diminishing the impact of local anomalies. Consequently, it reconstructs more realistic spatiotemporal traffic patterns, surpassing baseline models and achieving superior kriging performance even with limited observations.

(2) Effective spatiotemporal graph embedding techniques enhance model performance: spatiotemporal embedding, combining Eigenmaps graph embedding technology with sine-cosine function-based temporal embedding, builds richer feature representations and guides attention to discern spatiotemporal correlations, significantly boosting model efficacy.

(3) The design of spatiotemporal attention modules is crucial: The intricate temporal and spatial relationships within traffic networks prompted the creation of spatiotemporal attention modules. These modules direct the tensor mapped by spatiotemporal embedding to execute the same scaled dot-product attention operation in local subspaces, effectively capturing diverse spatiotemporal distribution patterns. Notably, the inclusion of Multi-head Spatial Interaction Attention notably enhances kriging accuracy.

Future enhancements for the proposed Kriformer could focus on two key aspects. Firstly,

exploring spatiotemporal matrices grounded on multi-relational graph structures, acknowledging the complexity and heterogeneity of node connections influenced by diverse relationships (Zhang and Kou, 2022). Secondly, Kriformer can refine spatiotemporal attention mechanisms based on multi-graphs to extract further benefits from multi-relational graph matrices.

**Acknowledgement**
The work described in this paper was supported by the National Natural Science Foundation, China (72025104, 72201214).

**Appendix A. Attention Mechanism**

The attention mechanism is a fundamental operation in our model designed to map a query and a set of key-value pairs to an output, where the query ($Q$), keys ($K$), values ($V$), and output are all vector representations. The output is a weighted sum of the values, where the weight assigned to each value is determined by the corresponding key and query. Each weight represents the strength of the relationship between the query and each key-value pair. In our study, we employ the scaled dot-product attention operation, originally introduced in the transformer model for attention mechanisms (see Figure 10). Scaled dot-product attention operates by computing weights through the dot product between the query and the values, providing advantageous traits such as spatial and temporal efficiency. The computation of scaled dot-product attention is outlined as follows:

$$\text{Attention}(Q, K, V) = V \cdot \text{Softmax}\left(\frac{QK^T}{\sqrt{d}}\right) \quad (10)$$

where $Q \in R^{n_q \times d}$, $K \in R^{n_k \times d}$ and $V \in R^{n_k \times d}$ $d$ are query, key, and value matrices respectively. In this context, $n_q$ and $n_k$ signify the input lengths. For a self-attention mechanism, $n_q = n_k$; however, in an interactional attention mechanism, $n_q \neq n_k$. The parameter $d$ represents the feature dimension. Additionally, to capture attention distributions among diverse patterns, the multi-head attention mechanism partitions the original input tensor into multiple smaller sub-tensors. This is typically done along the last dimension of the tensor, effectively splitting the information into different 'heads', and enabling attention calculation for each sub-tensor. The specific calculation method is detailed as follows:

$$\begin{cases} Q' = Split(Q) \in \mathbb{R}^{N_h \times n_q \times (d/N_h)} \\ K' = Split(K) \in \mathbb{R}^{N_h \times n_k \times (d/N_h)} \\ V' = Split(V) \in \mathbb{R}^{N_h \times n_k \times (d/N_h)} \\ \text{head}_h = \text{Softmax}\left(\frac{Q'_h(K'_h)^T}{\sqrt{d/N_h}}\right) V'_h \\ \text{Multi} - \text{head}(Q, K, V) = \text{concat}(head_h, \dots, head_{N_h}) \end{cases} \quad (11)$$

where $N_h$ denotes the number of sub-tensors or the number of heads, $h$ ($h = 1, \dots, N_h$) represents the sub-tensor index within this range.

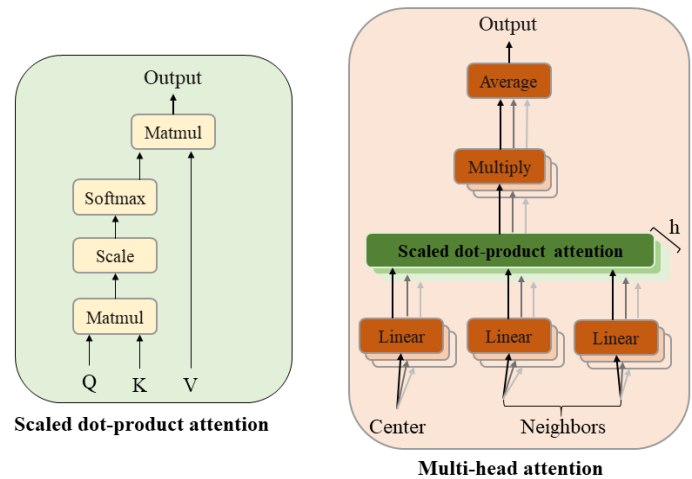

Figure 10: Scaled dot-product attention and Multi-head attention mechanism